\documentclass[10pt,twocolumn,letterpaper]{article}

\usepackage{cvpr}              
\usepackage[accsupp]{axessibility}  

\usepackage{graphicx}
\usepackage{amsmath}
\usepackage{amssymb}
\usepackage{booktabs}
\usepackage{multirow}
\usepackage{subcaption}
\usepackage{threeparttable}
\usepackage{array}
\newcolumntype{P}[1]{>{\centering\arraybackslash}p{#1}}

\makeatletter
\newcommand{\thickhline}{%
    \noalign {\ifnum 0=`}\fi \hrule height 1pt
    \futurelet \reserved@a \@xhline
}
\newcolumntype{"}{@{\hskip\tabcolsep\vrule width 1pt\hskip\tabcolsep}}
\makeatother

\usepackage[pagebackref,breaklinks,colorlinks]{hyperref}

\usepackage{float}

\usepackage[capitalize]{cleveref}
\crefname{section}{Sec.}{Secs.}
\Crefname{section}{Section}{Sections}
\Crefname{table}{Table}{Tables}
\crefname{table}{Tab.}{Tabs.}

\def\cvprPaperID{33} 
\def\confName{CVPR}
\def\confYear{2022}

\begin{document}

\title{GAF-NAU: Gramian Angular Field encoded Neighborhood Attention U-Net for Pixel-Wise Hyperspectral Image Classification}

\author{Sidike Paheding, Abel A. Reyes,   
Anush Kasaragod, 
Thomas Oommen \\
Michigan Technological University\\
Houghton, MI, USA\\
{\tt\small \{spahedin, areyesan, akasarag, toomme\}@mtu.edu}
}
\maketitle

\begin{abstract}

Hyperspectral image (HSI) classification is the most vibrant area of research in the hyperspectral community due to the rich spectral information contained in HSI can greatly aid in identifying objects of interest. However, inherent non-linearity between materials and the corresponding spectral profiles brings two major challenges in HSI classification: interclass similarity and intraclass variability. Many advanced deep learning methods have attempted to address these issues from the perspective of a region/patch-based approach, instead of a pixel-based alternate. 
However, the patch-based approaches hypothesize that neighborhood pixels of a target pixel in a fixed spatial window belong to the same class. And this assumption is not always true. To address this problem, we herein propose a new deep learning architecture, namely Gramian Angular Field encoded Neighborhood Attention U-Net (GAF-NAU), for pixel-based HSI classification. The proposed method does not require regions or patches centered around a raw target pixel to perform 2D-CNN based classification, instead, our approach transforms 1D pixel vector in HSI into 2D angular feature space using Gramian Angular Field (GAF) and then embed it to a new neighborhood attention network to suppress irrelevant angular feature while emphasizing on pertinent features useful for HSI classification task. Evaluation results on three publicly available HSI datasets demonstrate the superior performance of the proposed model. The source code available at \url{https://github.com/MAIN-Lab/GAF-NAU/}

\end{abstract}

\begin{figure}[ht]
  \includegraphics[scale=0.34]{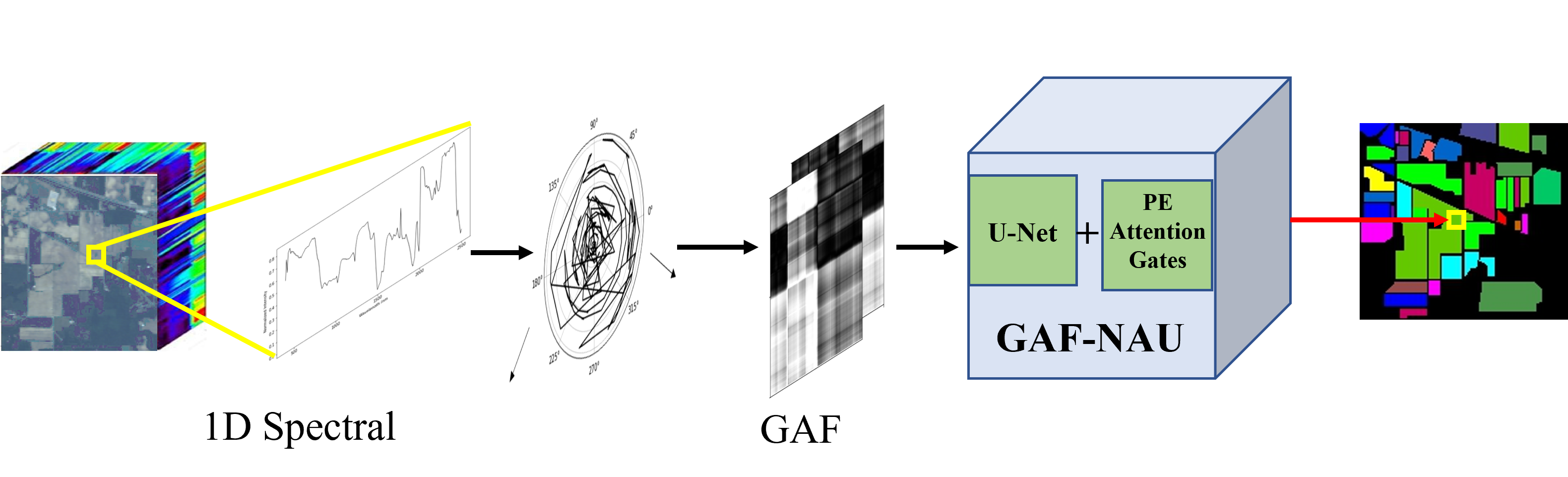}
  \caption{High-level schematic of the proposed methodology for pixel-wise HSI classification. GAF: Gramian Angular Field. GAF-NAU: Gramian Angular Field encoded Neighbor Attention U-Net.}
  \label{fig: 1}
\end{figure}

\section{Introduction}
\label{sec:intro}

Hyperspectral images (HSIs) contain abundant spectral bands/channels where each band measures the radiated energy from objects in narrow bandwidths. The detailed spectral information has several applications in fields of agriculture, forestry, urban and natural resources management pertaining to land use classification \cite{sidike2018progressively}, defense \cite{yuen2010introduction}, mineral mapping \cite{yokoya2016potential}, vegetation health analysis \cite{liang2015estimation}, etc. The challenging part, however, is to differentiate the variability in the spectral signatures of materials in a scene.

The conventional HSI classification involves the following steps: removing noisy bands (such as water absorption bands), data normalization, feature extraction, spectral and/or spatial feature based classification. Spectral-feature based classification approaches treat each HSI pixel as a 1D spectrum and then perform pixel-wise categorization, while spatial-feature based methods take into account the characteristics of pixels of its neighborhood in the spatial domain. Moreover, the spectral-spatial based classification frameworks utilize both spectral and spatial features during HSI classification. The limitation of the conventional classification approach is that feature engineering and classification are two separate tasks, and the hand-crafted features can hinder a classifier's performance if it is not carefully designed. Recent advances in machine learning, particularly, deep learning have shown great promises in achieving end-to-end HSI classification framework by automating the feature engineering process and has been able to produce models with demonstrated good performance \cite{audebert2019deep, li2019deep}.

Among various deep learning approaches for HSI classification, convolutional neural network (CNN) \cite{lecun1998gradient} has become the most popular one. CNN and its variants extract informative features from the original data via a series of hierarchical layers such as convolution, max pooling and fully connected layers. CNN has been employed to perform 1-D spectral, 2-D spectral-spatial, and 3-D spectral-spatial based HSI classification. \cite{hu2015deep,yu2017convolutional,audebert2019deep}. However, most recent works that involve deep CNN empathize on utilizing both spatial and spectral features to obtain benefits of 2D or 3D convolutional operations, instead of working on 1D pixel-vector itself due to the complex spectral property of HSI such as high-dimensionality and spectral correlation. Although 2D or 3D based CNN based frameworks have shown promises compared to 1D CNN methods, it assumes pixels within a fixed spatial size of neighborhood share similar spectral characteristics, which ignores that the possibility of pixels in the same neighborhood may represent a different class. This problem becomes more prominent if the spatial resolution is coarser. To address this issue, a robust deep learning framework capable of transforming 1D spectral vectors of hyperspectral data into 2D spectral feature matrices is needed to enable utilizing 2D CNN architectures, however, this type of approach has rarely been studied. 

In this research, we propose an effective pixel-wise HSI classification framework that represents a 1D spectral signature of a pixel vector as a 2D feature map in Gramian Angular Fields (GAF) and then embeds it to a deep network consisting of neighborhood attention gate, progressive expansion layer, and U-Net framework. Figure~\ref{fig: 1} shows a high-level schematic of our method. The proposed approach not only alleviates spectral correlation challenges as it considers CNN-based operations on angular fields, but also allows for integrating advanced CNN architectures into the HSI classification framework to account for interclass and intraclass spectral variations. The key contributions are summarized as follows:

\begin{itemize}
    \item To the best of our knowledge, this study is the first attempt to introduce the notion of 2D attention learning and 2D U-Net for a pixel-wise HSI classification task.
    \item The concept of attention gate with progressive expansion layer that connects encoder and decoder in U-Net framework is introduced, which utilizes attention from both higher and lower feature maps to highlight relevant features for better classification accuracy.
    \item We demonstrate the state-of-the-art classification performance on three benchmark HSI datasets by comparing existing CNN-based pixel-wise HSI classification frameworks.
\end{itemize}

In the rest of the paper, Section 2 reviews related pixel-wise HSI classification methods. Section 3 introduces the proposed approach, while the experimental analysis is presented in Section 4. Finally, we conclude the work in Section 5.

\section{Related Work}
\label{sec:RelatedW}

Spectral-based or pixel-based CNN models consider 1D spectral signature, denoted as $\textbf{x}_i \in R^B $, as an input, where $B$ represents the number of spectral bands in a given HSI data. In \cite{hu2015cnnHSI}, a simple 1D CNN architecture (i.e., one convolution, one pooling and one fully connected layer) is introduced for HSI classification by considering the spectral signature of each pixel as a 1D array, and yielded better accuracy than two-layer neural network and supper vector machine classifier. To alleviate the influence of strong correlation among HSI spectral bands, Gao et al. \cite{gao2018joint} reshaped the 1D spectral vectors of hyperspectral data into 2D spectral feature matrices and then adopted small convolution kernels with size of $3 \times 3$ or $1 \times 1$ to form convolutional layers. Recently, a new 1D CNN approach named PlasticNet \cite{jiang2021using} is introduced to identify plastic components from ATR-FTIR (attenuated total reflection-Fourier transform infrared spectroscopy) spectra. The PlasticNet receives 1D spectral signal collected from ATR-FTIR and represents it as Gramian angular fields (GAF) to form 2D matrix. This transformation allows for applying 2D CNN on 2D GAF, which produced higher accuracy in classifying mixed plastic waste comparing to 1D CNN. Their framework also employed a Piecewise Aggregate Approximation (PAA) \cite{keogh2001dimensionality} method to reduce the dimension of the input GAF matrices with the aim of ameliorating computational burden of 2D CNN. In \cite{wu2017RCNN}, recurrent layers are combined with convolutional layers to extract both contextual information and locally-invariant features from 1D spectral vector, the architecture achieved better accuracy than 1D and 2D CNN methods. Charmisha et al. \cite{charmisha2018dimensionally} performed a comparative analysis on the effect of a dimensionality reduction (DR) method, namely dynamic mode decomposition (DMD), for the performance of 1D CNN. The results indicate a comparable classification accuracy can be obtained with the reduced feature dimension by applying DMD. 
Zhu et al. \cite{zhu2018generative} improved the generalization capability of a CNN classifier by introducing generative adversarial network (GAN) in the HSI classification framework which tends to reduce overfitting issues associated with 1D-CNN.  Sidike et al. \cite{sidike2019dpen} proposed a pixel-wise deeper CNN classifier, namely deep Progressively Expanded Network (dPEN), which can extract pertinent features from raw data using very limited training data, and outperformed 1D-CNN and other popular machine learning classifiers. 

Recently, U-Net and its variants, as well as attention gates, have been employed and have shown promising results for HSI classification task, where a 2D patch-based semantic segmentation approach is typically implemented \cite{paul2022classification, hao2020geometry, mei2019spectral, hang2020hyperspectral, lin2021context}. To the best of our knowledge, there is no approach thus far that performs attention U-Net based pixel-wise HSI classification. In this work, we introduce a neighborhood attention U-Net scheme where attention gates encapsulate neighboring feature maps and progressive expansion layers. The proposed architecture encodes 1D pixel vector using Gramian Angular Summation Fields (GASF) and Gramian Angular Difference Fields (GADF) which are then fed to a neighborhood attention U-Net to perform pixel-wise classification. It is worth mentioning that U-Net based deep learning architectures are designed to output a 2D segmentation map in the last stage of the network, however, in the HSI classification task, it is expected to output a single value that indicates a class label. To address this issue, we also employ a majority voting scheme onto the segmentation map to calculate a class label, More details will be provided in Section \ref{sec:Methodology}.

\section{Proposed Methodology}
\label{sec:Methodology}

\begin{figure}[ht]
  \includegraphics[scale=0.19]{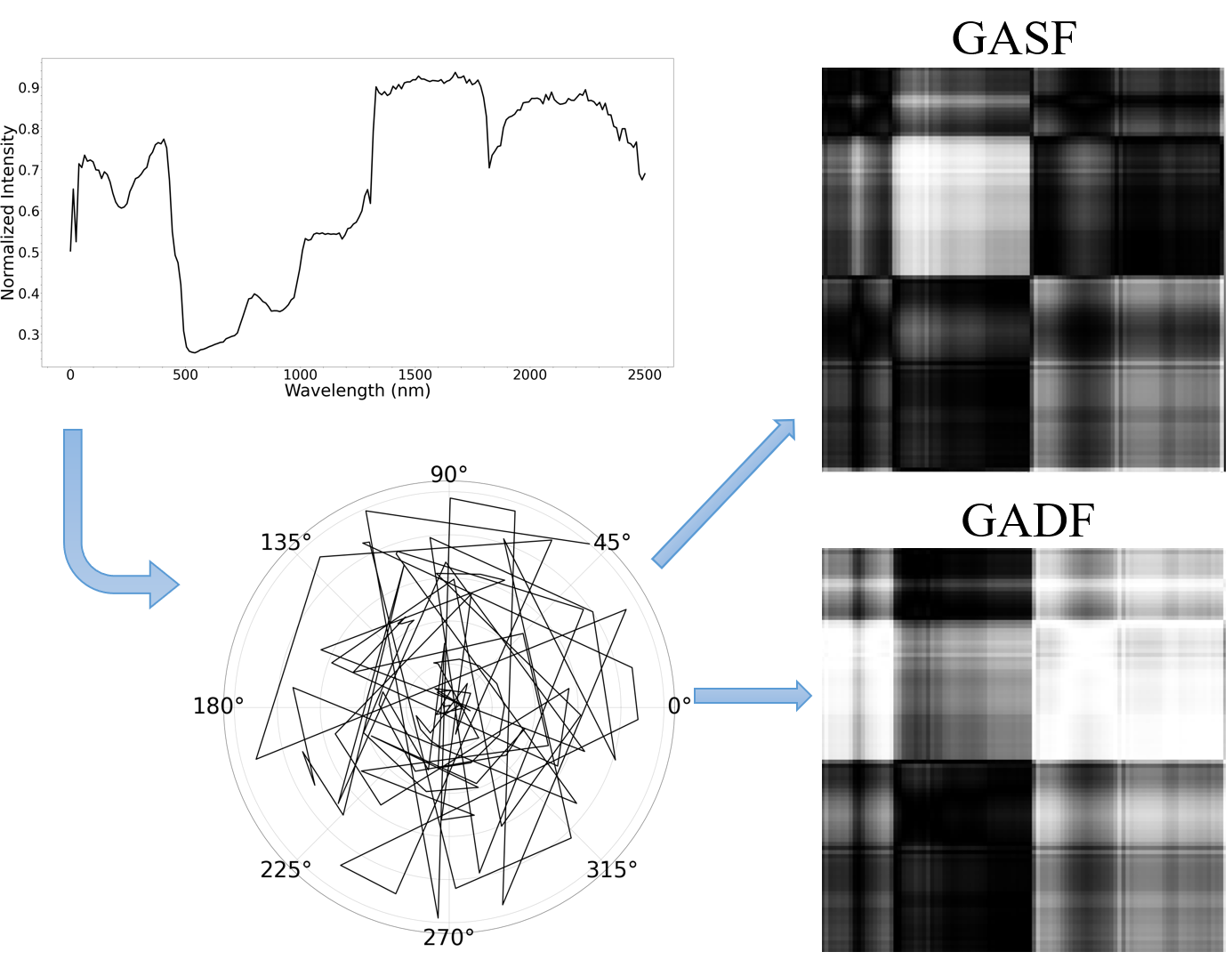}
  \caption{Conversion of 1D spectra to  the polar coordinate system and to GASF and GADF.}
  \label{fig:2}
\end{figure}

\subsection{Gramian Angular Fields (GAF)}
A HSI can be represented as a 3D cube as $\textbf{X} = \{{\textbf{x}_i}\}_{i=1}^{H \times W} \in R^{H\times W \times B}$ where $H$ and $W$ are the height and width of the spatial image, respectively. $B$ denotes the total number of spectral bands. $\textbf{x}_i \in R^B$ is the $i^{th}$ sample in the HSI cube with the $B$-dimension, which belongs to one of the available class $y_i \in \{1,2,\dots,C\}$. For 1D-CNN classification, each pixel is given as input as a 1D vector to a CNN architecture where 1D convolution and 1D pooling are applied.  In contrast, for 2D-CNN classification, either the 1D vector of spectra needs to be transformed into the 2D matrix, or a neighborhood window/patch of a center (target) pixel is treated as a sample. In this study, we employ GAF to transform a 1D pixel vector to the 2D matrix and then send it to a deeper network for prediction. Originally, GAF is used to encode time-series as images to capture correlation structures \cite{wang2015encoding} and use that output to process 2D-CNN. Similarly, GAF can be used to encode spectra as a 2D feature map to use as input to the 2D-CNN. In GAF, a 1D signal is represented in a polar coordinated system and the angles of each data point are converted into matrices using various operations as described in the following.  In the proposed framework, pixel vector is first normalized into [0, 1], expressed as

\begin{equation}
\Tilde{\textbf{x}}_i = \sum_{i=1}^N\frac{\textbf{x}_{i}-\textbf{X}_{min}}{\textbf{X}_{max} - \textbf{X}_{min}}
\label{eq:4}
\end{equation}
\noindent where $N$ (i.e., $H \times W$) is the total number of pixel in an HSI. Next, the $\Tilde{\textbf{x}}_i$ is represented in polar coordinate system by converting the normalized vector to angular cosine and radius with the equation below:

\begin{equation}
\phi_j = \arccos(\Tilde{\textbf{x}}_i^j), r_j = \frac{j}{B}, j=1,2,\dots,B 
\label{eq:5}
\end{equation}

\noindent where $\Tilde{\textbf{x}}_i^j$ represents the value of $j^{th}$ band in the normalized pixel vector $\Tilde{\textbf{x}}_i$. $\phi \in R^B$ is the angle vector and $r \in R^B$ is the radius vector. 

\begin{figure}[ht]
  \includegraphics[scale=0.26]{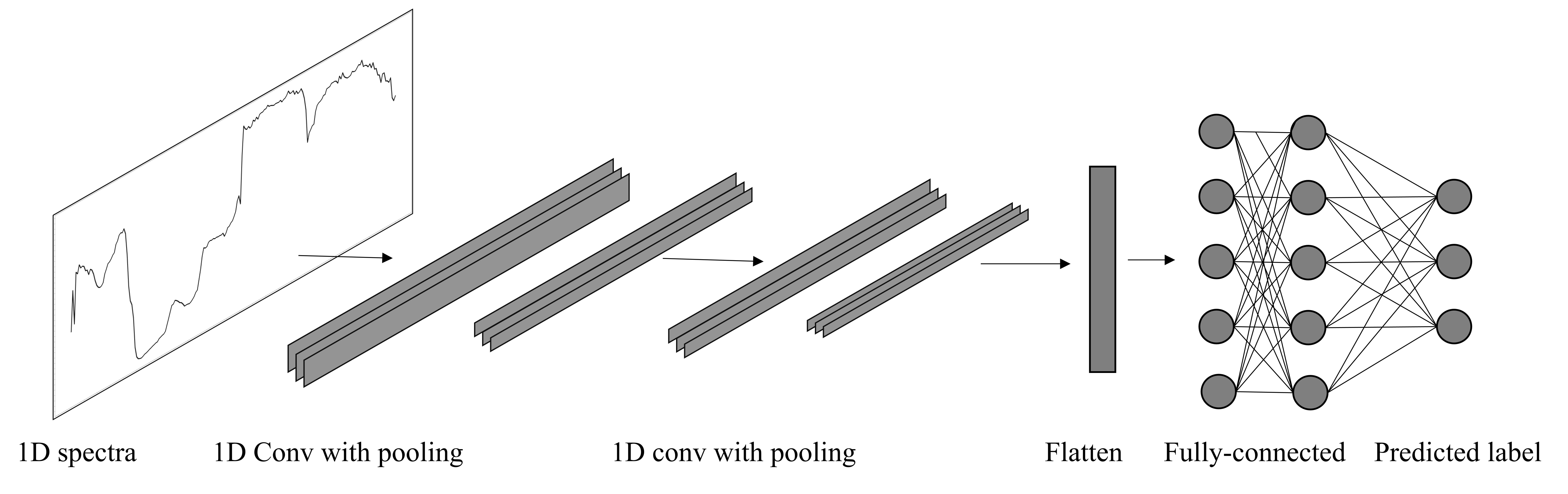}
  \caption{A typical architecture of 1D-CNN.}
  \label{fig:3}
\end{figure}

\begin{figure}[ht]
  \includegraphics[scale=0.253]{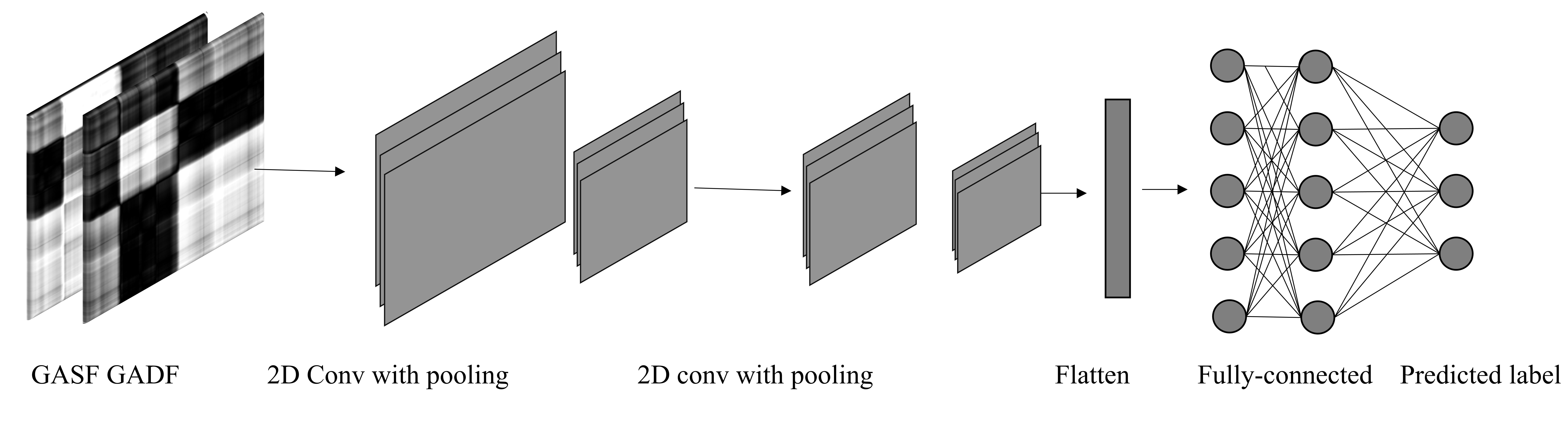}
  \caption{The architecture of GAF Encoded 2D-CNN.}
  \label{fig:4}
\end{figure}

There are two types of GAF: Gramian Angular Summation Fields (GASF) and Gramian Angular Difference Fields (GADF), which can respectively be calculated as

\begin{equation}
GASF = \cos(\phi_i+\phi_j) = \Tilde{\textbf{x}}^T\Tilde{\textbf{x}}-\sqrt{I-\Tilde{\textbf{x}}\textbf{x}^2}^T\sqrt{I-\Tilde{\textbf{x}}^2} 
\label{eq:6}
\end{equation}
\noindent and
\begin{equation}
GADF = \sin(\phi_i-\phi_j) = \sqrt{I-\Tilde{\textbf{x}}^2}^T\Tilde{\textbf{x}}-\Tilde{\textbf{x}}^T\sqrt{I-\Tilde{\textbf{x}}^2}
\end{equation}

\noindent where $i$ and $j$ denotes $i^{th}$ and $j^{th}$ spectral band, respectively. $I$ is a unit row vector of size $1\times B$ (i.e., the total number of spectral bands). 

\begin{figure*}[!ht]
    \centering
    \includegraphics[scale=0.38]{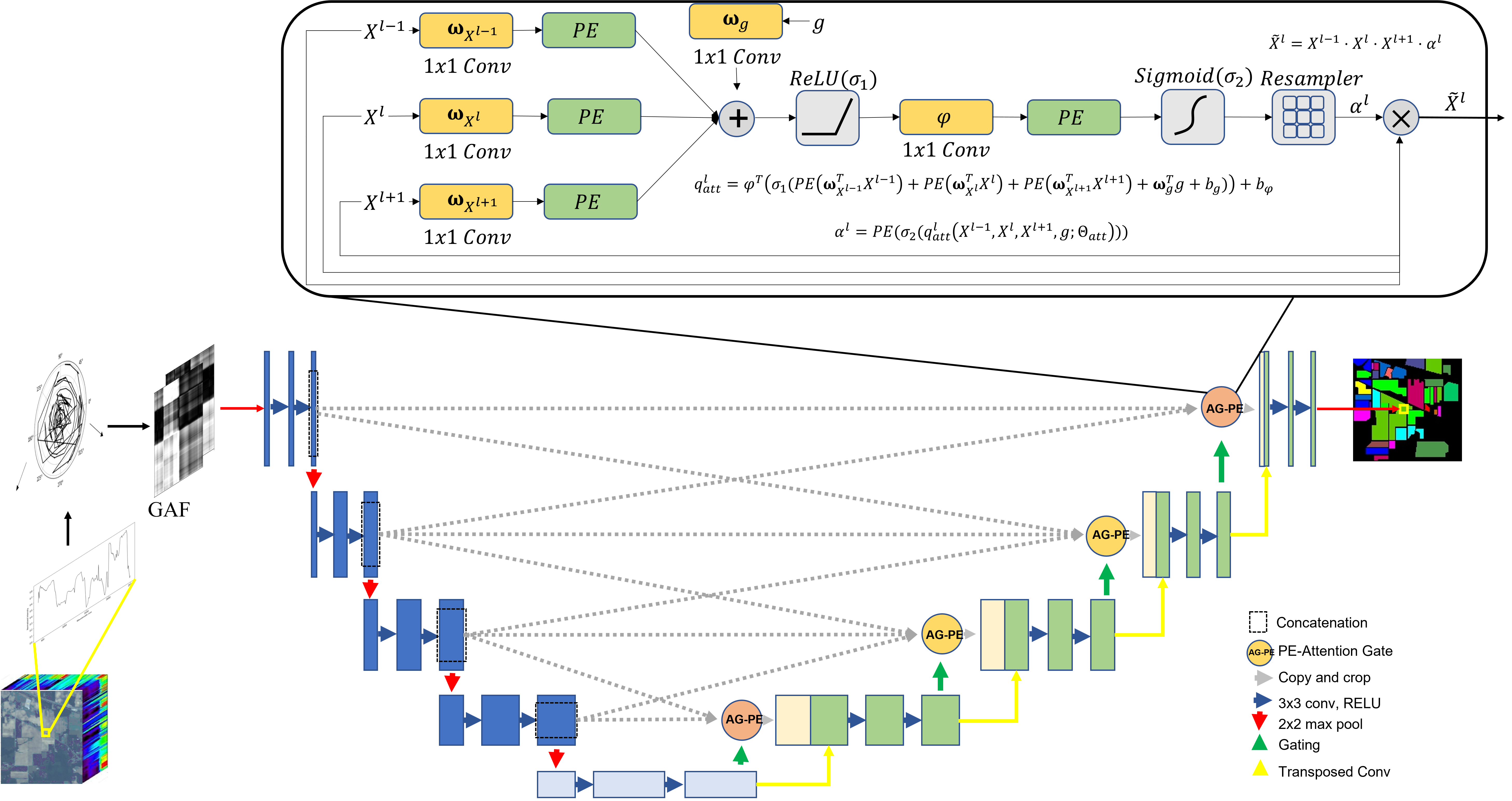}
    \caption{Schematic of the proposed GAF-NAU architecture, presented on the Indian Pines dataset. The first step consists of sending a HSI pixel to the GAF encoder module and then feed the output to the proposed AG-PE (Attention Gate with Progressive Expansion) embedded U-Net architecture.}
    \label{fig:UNET_PEN_Att}
\end{figure*}

The resulting GASF and GADF have a size $B \times B$, where $B$ is the number of the spectral band of HSI. The size of the GAF can be varied, and thus the corresponding prediction output is subject to change \cite{jiang2021using}. The conversion of 1D spectra to polar coordinates and to GASF and GADF is illustrated in Figure~\ref{fig:2}. The 1D-CNN architecture is shown in Figure~\ref{fig:3} and the GAF encoded 2D-CNN architecture is shown in Figure~\ref{fig:4}. 

\subsection{Proposed GAF-NAU Method}
We herein introduce a new HSI classification approach, namely Gramian Angular Field encoded Neighborhood Attention U-Net (GAF-NAU), which consists of neighbor attention mechanism, progressively expanded attention unit, and U-Net framework. The objective is to perform pixel-wise classification by using the encoded GAF feature map as an input to the proposed deep learning architecture to perform pixel-wise categorization. The proposed GAF-NAU architecture is depicted in Figure \ref{fig:UNET_PEN_Att} and illustrated in Section \ref{modelArchitecture}.

\begin{figure}[ht]
    \centering
    \includegraphics[scale=0.31]{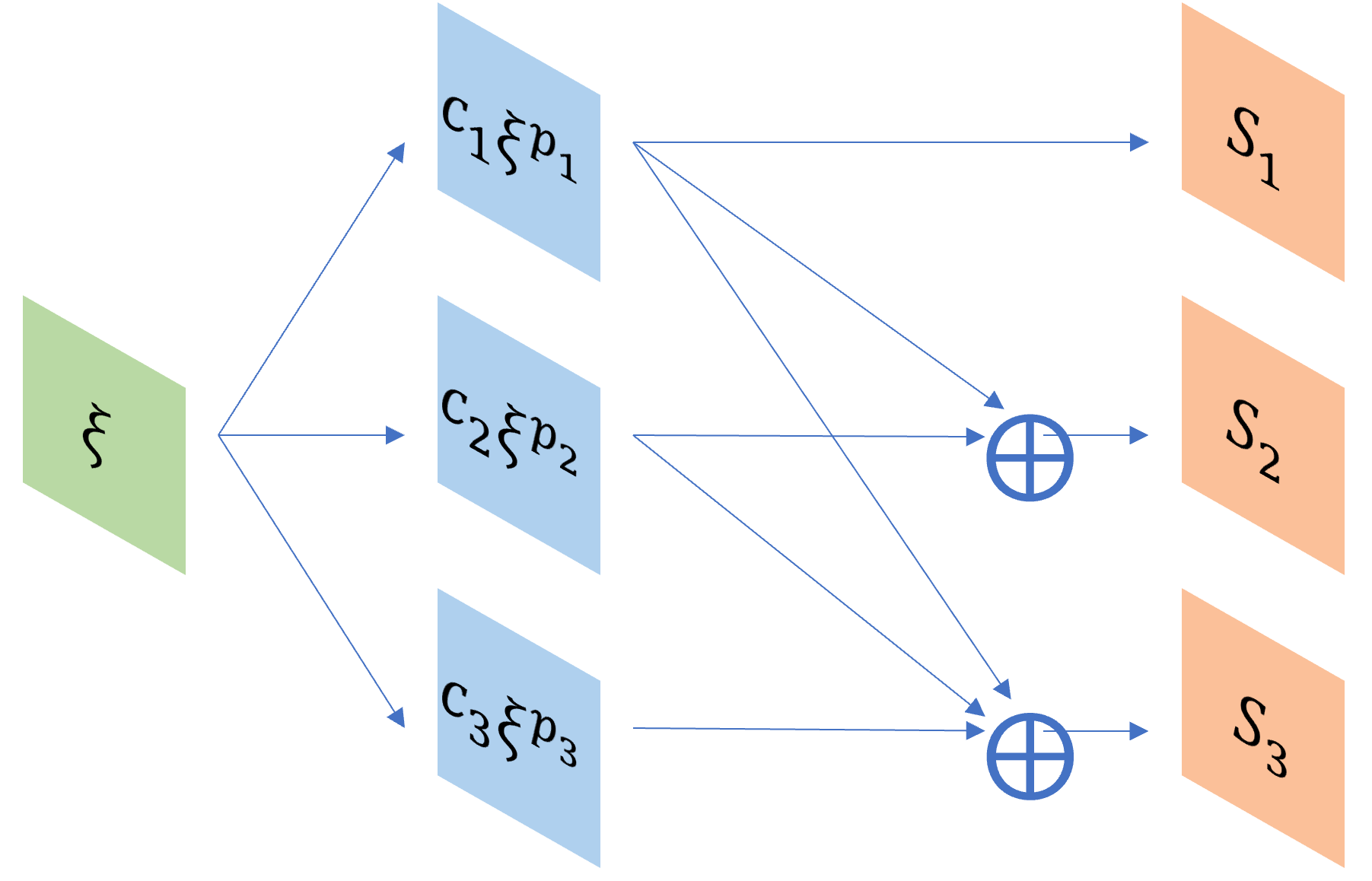}
    \caption{Illustration of the progressively expanded layer for a case of the Maclaurin series with three terms. $\xi$ represents an intermediate feature map in the network, $c_1$, $c_2$, and $c_3$ represent first three terms' coefficients of the Maclaurin series of a nonlinear function, respectively. And $p_1$, $p_2$, and $p_3$ represent the corresponding terms' powers. $S_k (k=1,2,3)$ are called progressively expanded feature maps.}
    \label{fig:PEL}
\end{figure}

\begin{figure*}[ht]
    \centering
    \includegraphics[scale=0.60]{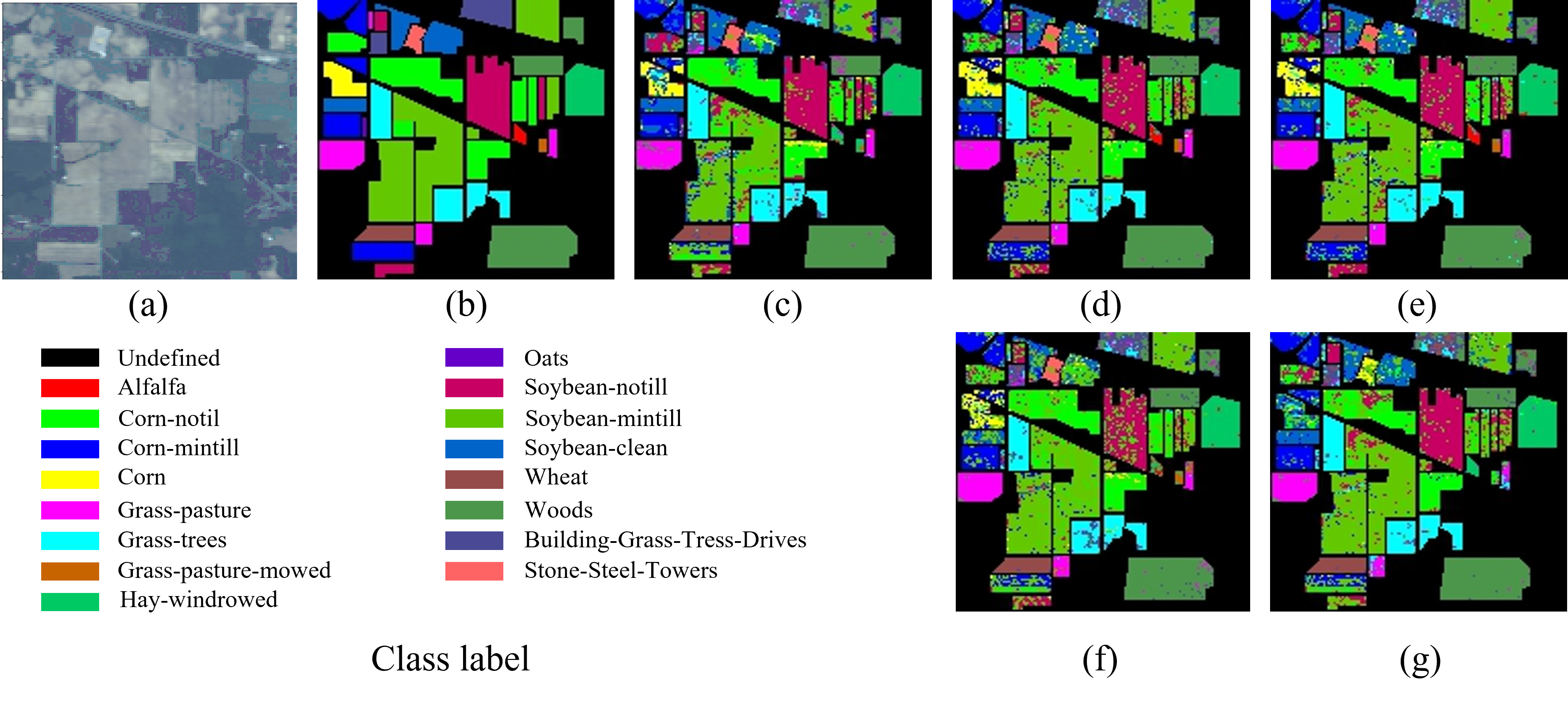}
    \caption{Indiana Pines hyperspectral dataset with classification maps. (a) True color composite of HSI, (b) Ground truth. Classification maps from (c) 1D-CNN, (d) dPEN, (e) Plastic-Net (f) GAF U-Net, (g) GAF-NAU.}
    \label{fig:5}
\end{figure*}

\subsubsection{Model Overview}
The main purpose of this research is to explore an alternate approach that enables deep 2D CNN-based methods to perform 1D pixel-wise classification in HSI, which has been rarely studied in the literature. In addition, we hypothesize that adding progressive expansion layer to the self-attention gating module can facilitate U-Net based deep network to learn more discriminate features than gating-based features propagated through the skip connections. The proposed method consists of two major steps: first, each pixel in HSI is passed through a GAF transformation which combines GASF and GADF to better capture spectral correlation. Second, the encoded GAF feature representation is fed to neighborhood attention U-Net to highlight salient features while suppressing irrelevant regions in GAF to better describe pixel category. Detailed architecture is described in the following section.

\subsubsection{Model Architecture} \label{modelArchitecture}
As shown in Figure \ref{fig:UNET_PEN_Att}, GAF-NAU contains three major components: the initial GAF transformation, Attention Gate with Progressive Expansion (AG-PE) block, and U-Net framework. In the AG-PE block, the PE layer is applied to $1 \times 1$ convolution (i.e., $\omega_X$). PE layer was originally developed for 1D pixel-wise multispectral image classification, and found to be effective in learning complex data structures by generating a nonlinear boundary that may better fit the nonlinear features \cite{sidike2019dpen}. In this study, we embed it in the self-attention gate and further extend it to perform 2D PE on each feature map. Figure \ref{fig:PEL} illustrates an example of a 2D PE layer. For a better explanation, let $\xi$ represents an input 2D feature map, then the $k^{th}$ progressively expanded feature map, denoted as $S_k$, can be expressed as

\begin{equation}
    S_k = \sum_{k=1}^{K} c_k\xi^{p_k} 
\end{equation}
\noindent where $c_k$ and $p_k$ are the coefficient and power of the $k^{th}$ term in the Maclaurin series of a nonlinear function (e.g., $\arctan$). $K$ denotes the total number of terms in the Maclaurin series is used. We set $K=2$ in this study. The coefficient and power of expanded terms are applied to input feature maps in an element-wise fashion (every node in the hidden layer). The purpose of introducing the PE layer in the self-attention gate is to contract robust feature maps using a nonlinear function with its corresponding Maclaurin series to better fit complex data structures. Since the PE layer does not contain trainable parameters, less model memory is consumed.

The AG-PE block is used to extract features not only from the same level of the encoder and decoder path in U-Net, but also from the neighbor's levels (upper and lower) as shown in Figure \ref{fig:UNET_PEN_Att}, and thereby we named our method as neighborhood attention. The output of the AG-PE block is element-wise multiplication of input feature-maps and attention coefficients (i.e., $\alpha \in [0,1]$): ${\widetilde{X}^l=X^{l-1} \cdot X^{l} \cdot X^{l+1}} \cdot \alpha^l $ where $X^{l}$ corresponds to the feature map in layer $l$. In Figure \ref{fig:UNET_PEN_Att}, the gating signal $g$ is used to determine focus regions, and it contains contextual information to trim lower-level feature responses \cite{oktay2018attention}. To obtain the gating coefficient, we employ additive attention \cite{oktay2018attention, bahdanau2014neural}, computed by 

\begin{equation}
    \begin{aligned}
q^l_{att} = \psi^T(\sigma_1(PE(\omega^T_{X^{l-1}} X^{l-1}) + PE(\omega^T_{X^{l}} X^{l}) +\\ PE(\omega^T_{X^{l+1}} X^{l+1})+ \omega^T_{g} g + b_g)) + b_\psi  
    \label{eq:5}
    \end{aligned} 
\end{equation}

\begin{equation}
    \alpha^l=PE(\sigma_2(q^l_{att}(X^{l-1},X^l, X^{l+1}, g; \Theta_{att})))
\end{equation}

\noindent where a ReLU and Sigmoid activation functions are represented as $\sigma_1$ and $\sigma_2$, respectively. $PE$ denotes progressive expansion operation. The set of parameters such as convolutions ($\omega_X$ and $\omega_g$) and bias terms ($b_g$ and $b_\psi$) are represented by $\Theta_{att}$. A more detailed explanation of similar terms can be found in \cite{oktay2018attention}. The proposed AG-PE block is incorporated into the modified U-Net architecture to highlight important features taken from the skip connections from the encoder to the decoder path as shown in Figure \ref{fig:UNET_PEN_Att}. The number of filter is set to 128 in the beginning and double it through the encoder path, and downsampling by 2 in the decoder path.  

The final layer of GAF-NAU produces a 2D output map where the values will be ranging from $[1 \ C]$ and $C$ denotes the total number of classes contained in the dataset. To determine the predicted label to the input pixel vector, Majority Voting (MV) strategy is adopted. following the MV rule, the majority class label from the output is considered as the final predicted label.

\begin{table*}[t!]
\small
\caption{Accuracy (in \%) comparison on the Indian Pines dataset.} 
\label{tab:IP_results}

\begin{center}  
    \begin{threeparttable}
    \begin{tabular}{|c|c|c|c|c|c|} \hline
    \rule[-1ex]{0pt}{1ex}  Metric & 1D-CNN \cite{hu2015cnnHSI}  &  dPEN \cite{sidike2019dpen}& Plastic-Net \cite{hu2015deep} & GAF U-Net & GAF-NAU \\\thickhline
    \rule[-1ex]{0pt}{1ex}  OA  & 74.39    & 77.62 & 73.00   & 75.67 & \textbf{81.07}   \\
    \rule[-1ex]{0pt}{1ex}  AA  & 76.36   & \textbf{81.28} & 63.29   & 71.69 & 74.67   \\
    \rule[-1ex]{0pt}{1ex}  $\kappa$  & 70.57   & 74.54 & 69.23   & 71.89 & \textbf{78.31}   \\ \hline
    
    \end{tabular}
    
    \begin{tablenotes}
      \footnotesize
      \item Note: The highest accuracy is highlighted in bold font.
    \end{tablenotes}
    
    \end{threeparttable}
\end{center}

\end{table*}

\begin{table*}[t!]
\small
\caption{Accuracy (in \%) comparison on the University of Pavia dataset.} 
\label{tab:UP_results}
\begin{center}       
\begin{tabular}{|c|c|c|c|c|c|}\hline
\rule[-1ex]{0pt}{1ex}  Metric & 1D-CNN \cite{hu2015cnnHSI}  & dPEN \cite{sidike2019dpen}& Plastic-Net \cite{hu2015deep} & GAF U-Net & GAF-NAU \\\thickhline
\rule[-1ex]{0pt}{1ex}  OA  & 90.17   & \textbf{91.21} & 89.70   & 89.70 &  91.12   \\
\rule[-1ex]{0pt}{1ex}  AA  & 89.37   & \textbf{90.54} & 87.76   & 87.70 & 90.49   \\
\rule[-1ex]{0pt}{1ex}  $\kappa$ & 86.85  & \textbf{88.29} & 86.29   & 86.25 & 88.09   \\\hline
\end{tabular}
\end{center}
\end{table*}

\begin{table*}[t!]
\small
\caption{Accuracy (in \%) comparison on the Salinas Valley dataset.} 
\label{tab:SV_results}
\begin{center}       
\begin{tabular}{|c|c|c|c|c|c|}\hline
\rule[-1ex]{0pt}{1ex}  Metric & 1D-CNN \cite{hu2015cnnHSI}  &  dPEN \cite{sidike2019dpen}& Plastic-Net \cite{hu2015deep} & GAF U-Net & GAF-NAU \\\thickhline
\rule[-1ex]{0pt}{1ex}  OA  & 91.43 & 92.53 & 90.69   & 93.82 & \textbf{94.59}   \\
\rule[-1ex]{0pt}{1ex}  AA  & 95.06  & 96.48 & 95.06   & 96.57 & \textbf{97.00}   \\
\rule[-1ex]{0pt}{1ex}  $\kappa$   & 90.45  & 91.68 & 89.63   & 93.11 & \textbf{93.97}   \\ \hline
\end{tabular}
\end{center}
\end{table*}

\section{Experimental Results}
\label{sec:Experiments}

\begin{figure}[ht]
    \centering
    \includegraphics[scale=0.62]{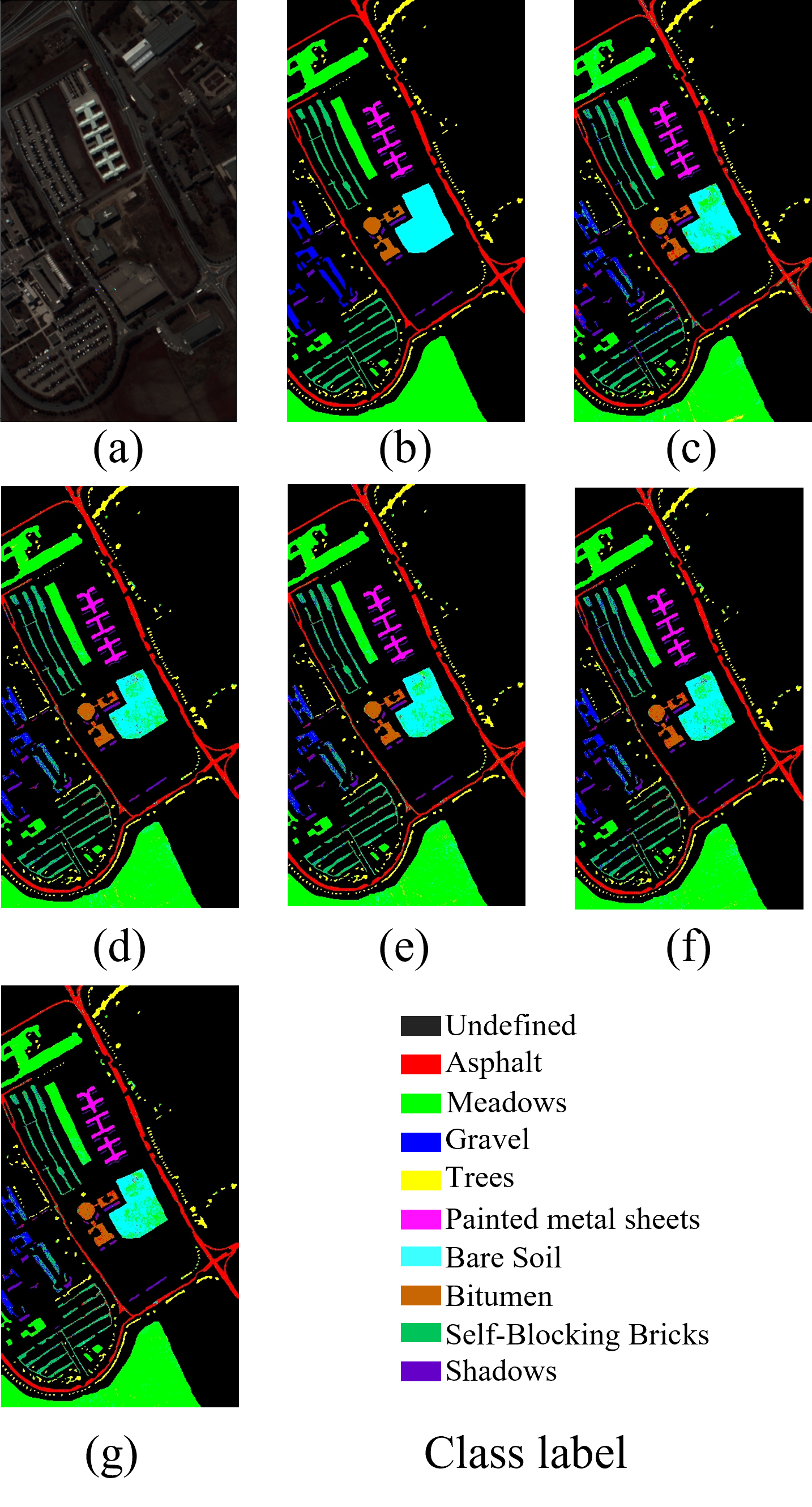}
    \caption{University of Pavia scene hyperspectral dataset with classification maps. (a) True color composite of HSI, (b) Ground truth. Classification maps from (c) 1D-CNN, (d) dPEN, (e) Plastic-Net (f) GAF U-Net, (g) GAF-NAU.}
    \label{fig:6}
\end{figure}

\begin{figure}[ht]
    \centering
    \includegraphics[scale=0.68]{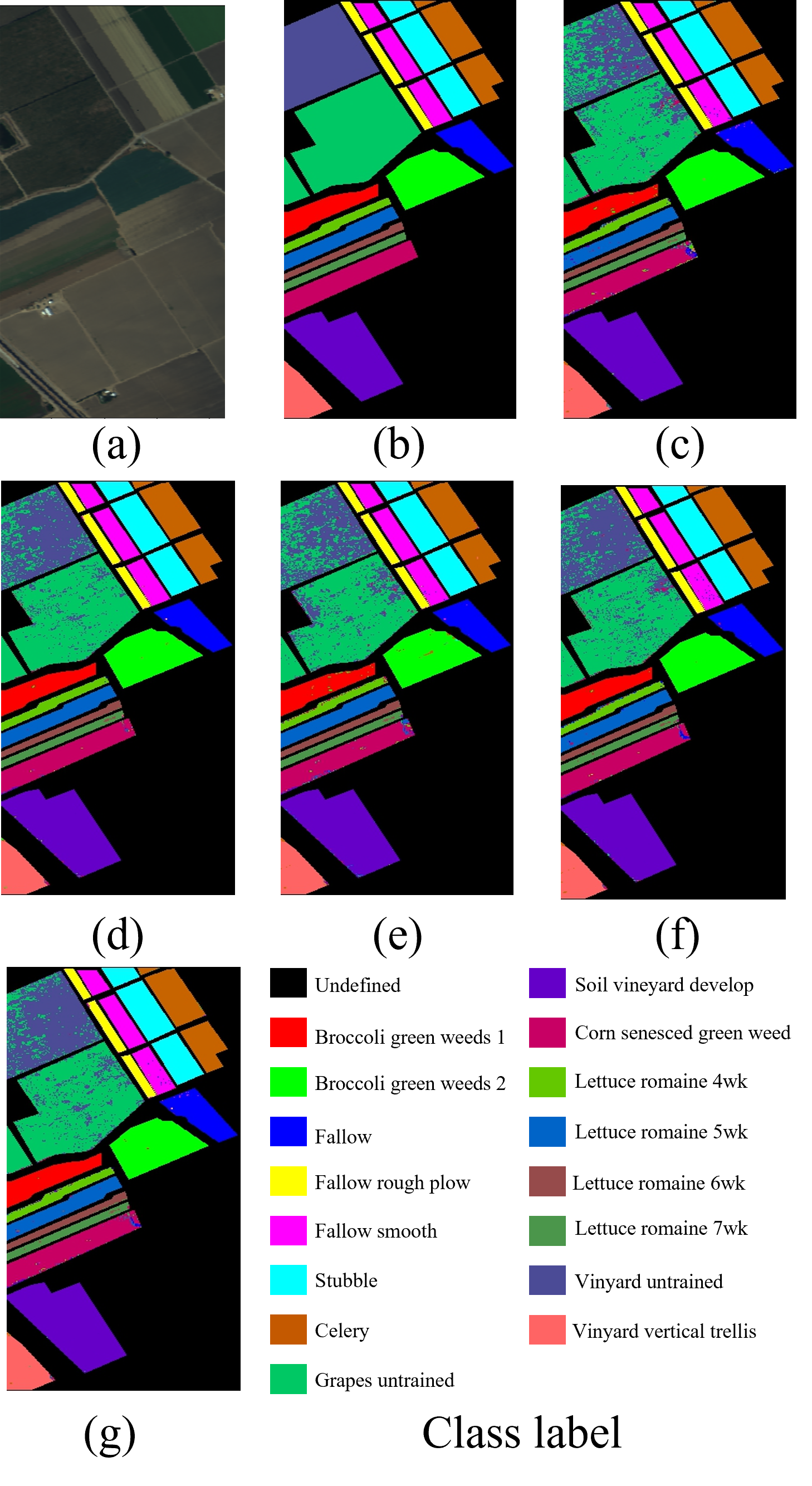}
    \caption{Salinas Valley hyperspectral dataset with classification maps. (a) True color composite of HSI, (b) Ground truth. Classification maps from (c) 1D-CNN, (d) dPEN, (e) Plastic-Net (f) GAF U-Net, (g) GAF-NAU.}
    \label{fig:7}
\end{figure}

\subsection{Datasets}
The datasets used in our experiments are Indian Pines (IP),
University of Pavia (UP), and Salinas Valley (SV), which are
publicly available \footnote{http://www.ehu.eus/ccwintco/index.php?title=Hyperspectral\_Remote\\ \_Sensing\_Scenes} and widely used for evaluation of HSI classification methods. The details of the datasets are described as follows.

\textit{Indian Pines (IP)}: Indian Pines dataset\cite{PURR1947} was acquired by an AVIRIS sensor covering the Indian Pines test area in the region of Northwestern Indiana. This dataset consists of $145 \times 145$ pixels and 224 spectral reflectance bands covering the range of 400–2500 nm. This scene is distributed as follows: two-thirds of agriculture presence and one-third of forest and/or natural perennial vegetation. There are a total of 16 classes, which are not all mutually exclusive, identified in the ground truth. By removing the water absorption region, 200 spectral bands are remained. Figure \ref{fig:5} (a) and (b) shows the pseudocolor image and the ground truth map of the IP dataset, respectively.

\textit{University of Pavia (UP)}: As part of the Pavia scenes, the University of Pavia HSI dataset was acquired by the Reflective Optics System Imaging Spectrometer (ROSIS) sensor over the city of Pavia, located in northern Italy. The size of the image is 610 $\times$ 610 with a total of 103 bands. There are nine types of land cover with a spatial resolution of 1.3 meters. Figure \ref{fig:6} (a) and (b) shows the pseudocolor image and the ground truth map of the UP dataset, respectively.

\textit{Salinas Valley (SV)}: The Salinas Valley HSI dataset was collected by the AVIRIS sensor over the Salinas Valley in California. The dataset comprises 512 by 217 pixels with a spatial resolution of 3.7-meter pixels, and contains a total of 220 spectral reflectance bands. In this dataset, the water absorption bands (20 in total) were also discarded, which left for 204 bands. The provided ground truth contains a total of 16 classes that cover regions of vegetables, bare soils, and vineyard fields. Figure \ref{fig:7} (a) and (b) shows the pseudocolor image and the corresponding ground truth map, respectively.

\subsection{Training Protocols}
Our proposed architecture is compared against other pixel-based deep learning architectures, including 1D-CNN \cite{hu2015cnnHSI}, dPEN \cite{sidike2019dpen} and Plastic-Net \cite{hu2015deep}. All the analyses have been placed under the same dataset split proportion for training ($10\%$), validation ($10\%$) and testing ($80\%$). To assess the performance of the different methods,  Overall Accuracy (OA), Average Accuracy (AA) and Kappa coefficient ($\kappa$) have been used as evaluation metrics. The proposed GAF-NAU and GAF U-Net use a fixed window size of 32 $\times$ 32 from the GAF matrix as an input for all the datasets. In the encoder path of GAF-NAU architecture, the number of filters in the convolutional layers is doubled after each down-sampling operation starting from 128 filters in the first convolutional layer. During the network training, the number of epochs is set to 150 epochs with a batch size of 64. And a learning rate scheduler was set, starting at $10^{-3}$ with constantly decreasing by a factor of $e^{-0.01}$ after each epoch. 










\subsection{Results}
Our proposed GAF-NAU architecture is verified with the IP, UP, and SV datasets, the corresponding results shown in Tables \ref{tab:IP_results}, \ref{tab:UP_results} and \ref{tab:SV_results}, respectively. It is noticeable that the proposed method outperforms the other state-of-the-art methods in the majority of experiments. For instance, GAF-NAU outperforms all competing methods in the IP dataset in terms of OA and $\kappa$. For the UP dataset, our method produces competitive performance compared to the dPEN algorithm. In the case of the SV dataset, our method again yields the best classification accuracy over OA, AA and $\kappa$.  It can be also observed that GAF-NAU consistently outperforms GAF U-Net for all three datasets, indicating the contribution of AG-PE block to the network performance. In terms of training speed, our proposed model takes longer time to train compared to other competing methods due to computational complexity of our algorithm. For instance, our method requires approximately 11.9 mins, 50.2 mins and 26.7 mins training time for IP, UP and SV datasets, respectively. In future work, we will explore strategies to improve its computational efficiency while maintaining good classification accuracy. 

The classification maps for IP, UP and SV datasets are demonstrated in Figures \ref{fig:5}, \ref{fig:6} and \ref{fig:7}, respectively. It can be visually observed that the classification maps produced from our GAF-NAU method tend to be less noisy and smoother in the majority of cases, which indicates better performance in handling interclass and/or intraclass variations. For instance, in Figure \ref{fig:5}, it can be seen that other competing methods incorrectly classify Soybean-notil as other classes, while our approach alleviates this problem at a certain level.

\subsection{Ablation Study}
In this section, we conduct a set of ablation studies to understand the contribution of various aspects of our proposed model. In Table \ref{tab:elements_results}, we evaluate the performance of three different models: 1) GAF-NAU, 2) GAF-NAU without the use of PE block in the AG, 3) GAF-NAU without the use of neighborhood AG-PE. As demonstrated in this table, the performance of the GAF-NAU model is decreased in the absence of any of those components, although the number of trainable parameters are reduced. 

Table \ref{tab:WS_results} presents OA on each dataset with different GAF matrix sizes, and it is observed that the GAF matrix size of $32 \times 32$ seems a reasonable choice considering OA obtained from all three datasets. There is a slight decrement of OA as the GAF matrix size increases for the IP dataset, while it is a no certain rule applied for both UP and SV datasets. However, it can be observed that the training time increases as the GAF matrix size increases. 

Furthermore, we verify the effect of increasing the training size and then evaluate the performance of our proposed model as shown in Table \ref{tab:TP_results}. As expected, the accuracy gradually increases as the number of training samples increases, which echos with typical deep learning models on the number of training samples requirement for achieving better performance.

\begin{table}[]
\small
\caption{Model performance (overall accuracy in \%) and trainable parameters by varying components of the proposed GAF-NAU architecture.} 
\label{tab:elements_results}
\begin{center}       
\begin{tabular}{|c|c|c|c|c|}
\hline
Model      & \begin{tabular}[c]{@{}c@{}} IP \end{tabular} & \begin{tabular}[c]{@{}c@{}} UP\end{tabular} & \begin{tabular}[c]{@{}c@{}}  SV\end{tabular} & \begin{tabular}[c]{@{}c@{}} \#Params \end{tabular} \\ \thickhline
GAF-NAU           & \textbf{81.07}                                              & \textbf{91.12}                                              & \textbf{94.59}                                              & 158.1 M                                                       \\
GAF-NAU w/o PE    & 79.91                                              & 90.63                                              & 93.98                                              & 154.0 M                                                        \\
GAF-NAU w/o AG-PE & 75.67                                              & 89.70                                              & 93.82                                              & 23.7 M                                                         \\ \hline

\end{tabular}
\end{center}
\end{table}

\begin{table}[]
\small
\caption{Model performance (overall accuracy in \%) and training time by varying GAF matrix size for three different datasets.} 
\label{tab:WS_results}
\begin{center}      
\begin{tabular}{|c|c|c|c|c|}
\hline
\begin{tabular}[c]{@{}c@{}}GAF \\ matrix size\end{tabular} & \begin{tabular}[c]{@{}c@{}} \\ IP\end{tabular} & \begin{tabular}[c]{@{}c@{}} \\ UP\end{tabular} & \begin{tabular}[c]{@{}c@{}} \\ SV\end{tabular} & \begin{tabular}[c]{@{}c@{}}Training\\ Time (s)\end{tabular} \\ \thickhline
GAF-NAU (16 x 16) & \textbf{82.01} & 89.10 & 93.44 & 2130.69 \\
GAF-NAU (32 x 32) & 81.07 & \textbf{91.12} & 94.59 & 3012.11 \\
GAF-NAU (48 x 48) & 79.73 & 90.99 & \textbf{94.79} & 4872.69 \\ \hline
\end{tabular}
\end{center}
\end{table}

\begin{table}[h]
\small
\caption{Model performance (overall accuracy in \%) by varying the fraction of training samples.} 
\label{tab:TP_results}
\begin{center}       
\begin{tabular}{|c|c|c|c|}\hline
\rule[-1ex]{0pt}{3.5ex}  Dataset & 10\% data & 15\% data & 20\% data  \\\thickhline
\rule[-1ex]{0pt}{1ex}  IP   & 81.07   & 84.91 & \textbf{85.98}  \\
\rule[-1ex]{0pt}{1ex}  UP  & 91.12   & 91.54 & \textbf{91.77} \\
\rule[-1ex]{0pt}{1ex}  SV  & 94.59   & 95.31 & \textbf{95.46} \\\hline
\end{tabular}
\end{center}
\end{table}

\section{Conclusion}
We proposed a new pixel-wise hyperspectral image classification framework, which we refer to as Gramian Angular Field encoded Neighborhood Attention U-Net (GAF-NAU). It combines multiple unique components, including GAF encoding, neighborhood attention gate, and progressive expansion layer, to ensure good feature representation. The experimental evaluation on three HSI datasets confirms the efficacy of the proposed network. Moreover, an ablation study was performed to investigate characteristics of the proposed GAF-NAU, which we found that 1) GAF matrix size affects classification output, and 2) Attention Gate with Progressive Expansion (AG-PE) has a noticeable contribution to the classification performance. It is anticipated that further gains in accuracy of GAF-NAU may be obtained by more detailed tuning of hyperparameters and proper combination of different network components, we plan to conduct such experiments in our future work. 

\section*{Acknowledgement}
Research reported in this publication was supported in part by funding provided by the National Aeronautics and Space Administration (NASA), under award number 80NSSC20M0124, Michigan Space Grant Consortium (MSGC).

{\small
\bibliographystyle{ieee_fullname}
\bibliography{gafNAU}
}

\end{document}